# Deepfake detectors are DUMB: A benchmark to assess adversarial training robustness under transferability constraints

**Adrian SERRANO**
Thales

**Erwan UMLIL**
Thales
erwan.umlil@thalesgroup.com

**Ronan THOMAS**
Thales
ronan.thomas@thalesgroup.com

## Abstract

Deepfake detection systems deployed in real-world environments are subject to adversaries capable of crafting imperceptible perturbations that degrade model performance. While adversarial training is a widely adopted defense, its effectiveness under realistic conditions - where attackers operate with limited knowledge and mismatched data distributions - remains underexplored. In this work, we extend the DUMB - Dataset soUrces, Model architecture and Balance - and DUMBer methodology to deepfake detection. We evaluate detectors robustness against adversarial attacks under transferability constraints and cross-dataset configuration to extract real-world insights. Our study spans five state-of-the-art detectors (RECCE, SRM, XCeption, UCF, SPSL), three attacks (PGD, FGSM, FPBA), and two datasets (FaceForensics++ and Celeb-DF-V2). We analyze both attacker and defender perspectives mapping results to mismatch scenarios. Experiments show that adversarial training strategies reinforce robustness in the in-distribution cases but can also degrade it under cross-dataset configuration depending on the strategy adopted. These findings highlight the need for case-aware defense strategies in real-world applications exposed to adversarial attacks.

## 1 Introduction

**Adversarial examples**, inputs perturbed by small, carefully crafted noise, pose a critical threat to machine learning systems. Their transferability across models enables black-box attacks, making defenses challenging in practice. While adversarial training (AT) is widely adopted to improve robustness, its effectiveness in real-world scenarios, where attackers and defenders operate with different architectures and data distributions, remains underexplored.

**Deepfakes** - DeepfakeBench [1] introduced a classification of deepfake detectors into three categories according to the information leveraged: naive, spatial and frequential. Deepfake detection is particularly vulnerable due to its reliance on subtle spatial and frequency artifacts, which adversarial perturbations can easily disrupt.

**Adversarial training (AT)** - augmenting training data with adversarial examples, remains one of the most studied **defenses** to improve robustness against such attacks. Yet, AT is studied under white-box assumptions ignoring realistic conditions where attackers operate with different architectures or datasets. This gap limits our understanding of AT's practical value.

**The DUMB framework** [2] formalized attacker-defender mismatch scenarios along dataset, model, and class balance axes, while DUMBer [3] extended this to adversarially trained models, revealing significant gaps in robustness under transfer-based attacks. To address this, we build on the DUMB/DUMBer framework and propose a systematic evaluation of adversarial robustness for deepfake detection under realistic mismatch scenarios. We study the following questions: (i) How effective are adversarial attacks when the attacker and defender differ in architecture or dataset? (ii) Does adversarial training maintain clean accuracy while improving robustness? (iii) Does increasing diversity in adversarial training (multiple attacks and surrogates) enhance defense performance?

In this paper, we investigate how different adversarial training strategies impact robustness, using deepfake detection as a concrete application. Specifically, we explore the effects of varying the diversity of attacks and surrogate models on adversarial training. Our experiments involve representative deepfake detector architectures, various training strategies, and evaluations on both in-distribution and cross-dataset settings.

## 2 Related works

### 2.1 Deepfake detection

Deepfakes refer to images or videos in which a person's face has been synthetically altered or generated through advanced AI techniques. The rapid progress of generative adversarial networks (GANs) [4] and diffusion-based models [5], [6] has enabled the creation of highly realistic deepfakes, which are commonly classified into four categories: face swapping, face editing, entire face synthesis, and face reenactment. Face swapping replaces the identity of a person in a video with another, face editing modifies facial attributes such as age, expression, or gender, entire face synthesis generates non-existent faces from scratch and face reenactment transfers the facial movements or expressions of a source to a target identity. Strategies in the literature to detect these vary: some methods target visual artifacts in the pixel domain, whereas others exploit inconsistencies in the frequency domain [7]. However, some studies have shown that deepfake detectors are vulnerable to both spatial and frequency-domain perturbations [8], [9].

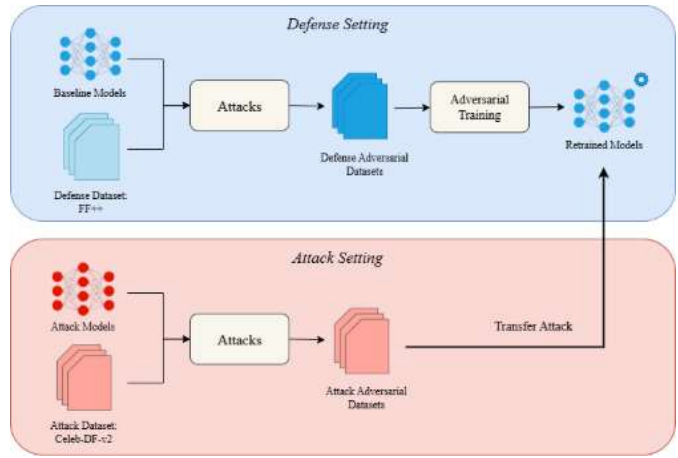

*Figure 1: Attack and defense settings*

## 2.2 Adversarial attacks

**Adversarial attacks** aim to subtly perturb input images to cause incorrect model predictions, often without perceptible changes to human observers. **White-box attacks** assume full access to a model's architecture, parameters, and gradients, enabling precise optimization of perturbations. Classic examples include Projected Gradient Descent (PGD) [10] and Fast Gradient Sign Method (FGSM) [11]. Some variants of these attacks turn out to be very effective. However, in practical applications, target models rarely expose such internal information, rendering white-box attacks largely unrealistic. In contrast, **black-box attacks** assume limited or no knowledge of the target model. These attacks are typically divided into three categories [12]: decision-based, score-based, and transfer-based attacks. **Decision-based** [13] **and score-based** [14] attacks rely on extensive queries to the target model to iteratively optimize perturbations, which are often not feasible in real-world scenarios. **Transfer-based** [10,11,15] attacks generate adversarial examples on surrogate models and apply them to the target, exploiting adversarial transferability—perturbations crafted for one model often succeed against others. Some of them are particularly threatening for Deepfake detection such as Frequency-based Post-train Bayesian Attack (FPBA) [16], Fakepolisher [17], TraceEvader [18] and S2I-FGSM [19]. Requiring no access to target parameters and only a single application, transfer-based attacks closely reflect realistic threat scenarios [20,21,22], which is the focus of our study.

## 2.3 Adversarial training strategies

**Adversarial training**, which incorporates adversarial examples into the training process, remains the most effective defense for improving model robustness. In [23], the authors propose a detailed review of AT methods. The main challenge of adversarial training lies in enhancing robustness to adversarial examples without significantly compromising accuracy on clean data. Prior work [24] highlights the inherent trade-off between these objectives. To address this, TRADES [25] introduces a regularization term that explicitly balances natural accuracy and adversarial robustness. Another limitation of standard AT is its reliance on a single attack, such as FGSM, PGD, or one of their variants, with evaluations typically restricted to white-box settings. While this can provide robustness against the specific attack used during training, the approach is susceptible to catastrophic overfitting [26], which reduces the resilience of the retrained models to unseen attacks. To overcome this issue, several methods have been proposed: Adaptive [27] or Curriculum [28] vary the perturbation budget, Ensemble [29] leveraged both PGD and FGSM in the training.

The DUMB [2] and DUMBer [3] frameworks offer systematic evaluation protocols for transferability but have not been applied yet to deepfake detection. Moreover, limited research has assessed adversarial training for deepfake detectors under transfer-based attacks and dataset shifts. Our work addresses this underexplored area, providing insights into deepfake detection under realistic attacks and dataset-shift scenarios.

## 3 Method

### 3.1 Adoption of the DUMB taxonomy

We extend the DUMB taxonomy [2] to assess the transferability of attacks, which formalizes attacker-defender mismatch scenarios along three axes: **D**ataset so**U**rce, **M**odel architecture, and class **B**alance. In this work, we focus on the most practically relevant and interpretable subset by adopting a balanced class configuration (B: matched), which simplifies the DUMB notation with four cases:

- C1 (White box): Attacker and defender share the same dataset and model architecture.

- C3 (Cross-model): Same dataset, different model architectures.

- C5 (Cross-dataset): Different datasets, same model architecture.

- C7 (Cross-dataset & cross-model): Both dataset and model architecture differ.

The attacker-defender approaches are summarized in Fig 1. The defense setting assesses the robustness and value of AT strategies, while the attack setting allows us to reason about real-world conditions where attackers rarely share the same data distribution as defenders.

### 3.2 Model selection

To ensure a comprehensive evaluation of adversarial robustness in deepfake detection, we selected five state-of-the-art models that collectively represent the diversity of current detection strategies, based on their performances on recognized benchmarks and on code availability for reproducibility purposes. Based on DeepfakeBench [1] and their coverage of both spatial and frequency-based detectors, we chose Xception [30] (2017), UCF [31] (2023), RECCE [32] (2022), SPSL [33] (2021) and SRM [34] (2021). Recent architectures [35-41] are identified as promising to extend our work.

**Xception** [30] is a widely adopted convolutional neural network architecture that leverages depth wise separable convolutions and has become a standard baseline for image forensics and deepfake detection due to its efficiency and strong performance on large-scale datasets.

**UCF** [31] Uncovering Common Features introduces a disentanglement that separates image features into content features, method-specific forgery components and common forgery components. This split improves generalization across unseen manipulation methods using a multi-task classification loss and a contrastive regularization loss.

**RECCE** [32] employs a dual-task strategy that combines image reconstruction with classification to improve deepfake detection. The reconstruction branch acts as a regularizer, forcing the shared encoder to learn fine-grained facial representations, while the classification branch leverages these features for forgery detection. This integrated design enables RECCE to capture both pixel-level anomalies and semantic inconsistencies.

**SPSL** [33] analyzes the **phase spectrum distortions in the frequency domain** that are created by up sampling and blending operations during face forgery. Instead of relying on deep semantic features, SPSL uses a shallow CNN to capture local texture and phase irregularities.

**SRM** [34] is a frequential approach that leverages learnable filters to capture high-frequency noise residuals that are characterized by forgery creation. These features are fed into a two-stream network (RGB+SRM), so the model effectively generalizes across diverse manipulation methods and compression levels.

### 3.3 Adversarial attacks

We consider three adversarial attacks to cover both spatial and frequency domains which are essential in the deepfake detection task. FGSM [11] is a single-step white-box attack that perturbs the input image pixels along the gradient sign direction. PGD [10] is an iterative stronger variant of FGSM. As gradient-based attacks, FGSM and PGD ensure compatibility with existing literature. FPBA is an attack designed to combine both spatial and frequency artifacts, which has been specifically designed to fool deepfake detectors. These three attacks provide a representative and diverse set of threat models relevant to both literature and real-world scenarios.

### 3.4 Adversarial training strategies

We investigate eleven adversarial training strategies. We define **single** (FGSM, PGD, FPBA), **dual** (PGD+FGSM, PGD+FPBA, and FGSM+FPBA) and **triple** (FGSM+PGD+FPBA) attacks AT strategy where we retrain the model with adversarial samples from a single, two or all three attacks using the same architecture. The **Ensemble** configuration (triple case) aims to increase robustness through attack diversity. Like Ensemble, we define **Surrogate** AT strategy where we use adversarial examples built by different model architectures. In this case, we define three strategies: PGD_surr, FGSM_surr, FPBA_surr, where we craft examples using one attack but multiple surrogates. Finally, we define Ens_Surr (i.e PGD_surr+FGSM_surr+FPBA_surr) where we bring the most diversity to the retraining by crafting examples with the three attacks and all surrogates.

### 3.5  Evaluation protocol

The deepfake detection task is formulated as a binary classification problem, where a model learns to distinguish genuine images from deepfakes based on the provided images. We use the AUC to evaluate models' nominal performance after AT. The Attack Success Rate (ASR) is used to measure the fraction of attacked samples misclassified by the model after attack. The Attack Mitigation Rate (AMR) defined in Eq. 1 corresponds to the relative reduction in ASR after adversarial training. ASR and AMR, define in [3] are used to evaluate models' robustness.

$$AMR = \frac{ASR_{orig} - ASR_{AT}}{ASR_{orig}} \quad (Eq. 1)$$

To comprehensively assess robustness, each retrained model is evaluated on 32 test sets: 16 from FF++ and 16 from Celeb-DF-V2. For each dataset, the test sets consist of 1 clean set and 15 adversarial sets generated using 3 attacks (PGD, FGSM, FPBA) with 5 nominal models trained on FF++ as surrogates. For each test set, we evaluate 12 training strategies (1 original + 11 adversarial) across 5 model architectures, yielding a total of 32 × 12 × 5 = 1920 evaluations. We present our results as an average across the detectors. Let $\mathcal{M} = \{M_i \mid i \in [1,5]\}$ represent our deepfake detection models, $\mathcal{A} = \{A_k \mid k \in [1,3]\}$ the set of adversarial attacks. Let $\mathcal{D} = \{D_j \mid j \in [1,2]\}$ represent our test datasets. We note $N_{success}^{(M_i, A_k)}$ the number of successful adversarial attacks on model $M_i$ using attack $A_k$ on a test dataset. Similarly, we note $N_{success}^{(M_t \leftarrow M_s, A_k)}$ the number of successful attacks on target model $M_t$ using adversarial examples generated by surrogate model $M_s$ with attack $A_k$. Let $N_{total}^{(M_i, A_k)}$ and $N_{total}^{(M_t \leftarrow M_s, A_k)}$ represent the total number of attacked samples for model $M_i$ using attack $A_k$, and the total number of attacked samples in a cross-model setup. Eq. 2 and Eq. 3 give the ASR on C1 and C3 cases, where models are trained and tested on FF++. The ASR on C5 and C7 cases correspond to Eq. 2, and Eq.3 with dataset $D_2$, and models trained on **FF++** and inference done on **Celeb-DF-V2**. They are provided in Appendix.

$$\text{ASR}_{C1}^{(D_1)} = \frac{1}{|\mathcal{M}| \cdot |\mathcal{A}|} \sum_{A_k \in \mathcal{A}} \sum_{M_i \in \mathcal{M}} \frac{N_{success}^{(M_i, A_k, D_1)}}{N_{total}^{(M_i, A_k, D_1)}} \quad (Eq. 2)$$

$$\text{ASR}_{C3}^{(D_1)} = \frac{1}{|\mathcal{M}|(|\mathcal{M}| - 1) \cdot |\mathcal{A}|} \sum_{A_k \in \mathcal{A}} \sum_{M_t \in \mathcal{M}} \sum_{M_s \in \mathcal{M}, M_s \neq M_t} \frac{N_{success}^{(M_t \leftarrow M_s, A_k, D_1)}}{N_{total}^{(M_t \leftarrow M_s, A_k, D_1)}} \quad (Eq. 3)$$

This setup enables a comprehensive assessment of model robustness across multiple attacks, surrogate models, and datasets, simulating different levels of knowledge available to the attacker.

### 3.6 Implementation

We gather here all the details related to the experiment's implementation.

**Datasets.** We adopt two widely used deepfake datasets to simulate both in-distribution and out-of-distribution scenarios. **FaceForensics++** [42] (FF++_c23) is used for model training and in-distribution evaluation. It contains real and fake images generated by four forgery methods, three face swapping and a face reenactment: DeepFakes, Face2Face, FaceSwap, and NeuralTextures. We split FF++ into training, validation and test sets using ratios of 70%, 10% and 20%. We apply video-level split to ensure that the sets are built without identity leakage to avoid contamination. **Celeb-DF-V2** [43] is used for out-of-distribution evaluation, simulating a real-world scenario where attackers have no access to the training data distribution. It leverages a single face-swapping method distinct from the ones used in FF++. This distribution shift results in a more challenging classification task. We use the preprocessed Celeb-DF-V2 version provided by DeepfakeBench [1].

**Detectors**. We replicate the training methodology from [1] in our detectors' training: models are trained from scratch using Adam optimizer, data augmentation [1], a batch size of 32 to minimize the binary cross-entropy. Complete training configurations from each model can be find in [1]'s GitHub.

**Attacks.** As attacks must remain imperceptible to the human eye, we conduct a qualitative study to determine the perturbation budget. We constrain all attacks by a $\ell\infty$ perturbation budget of 9/255. We use the PGD and FGSM implementation provided by the TorchAttacks library [44]. We run PGD with alpha = 4/255, 4 steps and random start. The number of steps used to run PGD was determined after analyzing how many steps PGD required to reach the maximum perturbation budget. We implement FPBA following the pseudo-code description provided. We use a standard gradient descent to sample the appended model's theta', use K=3 appended models, a learning rate of 0.001, N=5 random spectrum transformations per attack step, and 4 iterations. We first train our architectures on clean data to obtain a baseline on nominal conditions. We reproduce DUMB and DUMBER configuration, where all training strategies are performed from scratch rather than fine-tuning, and build our training sets using 80% clean samples and 20% adversarial samples. Adversarial images are generated from the training set and paired with their original images to maintain consistency across retraining strategies. When multiple attacks or surrogates are included, the 20% adversarial subset is divided equally among them.

## 4 Experiments

### 4.1 Attack effectiveness

We first focus on understanding the practical threats that adversarial attacks represent in the deepfake detection field. We systematically evaluate the ASR of three representative attacks - FGSM, PGD, FPBA - across DUMB scenarios: white-box (C1), cross-model (C3), cross-dataset but same model (C5) and cross-dataset & cross-model (C7). Results are summarized in Table 1.

PGD consistently achieved the highest ASR in all settings, with near-perfect success in the white-box scenario (C1: 99.6%) and substantial transferability in cross-model (C3: 89.8%) and cross-dataset (C5: 65.1%; C7: 66.7%) cases. FGSM, while less effective than PGD, still demonstrated significant transferability, particularly in white-box and cross-model settings (C1: 71.7%, C3: 56.5%). FPBA, designed to exploit both spatial and frequency artifacts, was especially effective against frequency-based detectors, but its overall transferability was lower than PGD. The detailed ASR performances per-model and per-attack can be found in Appendix.

These results confirm that transfer-based adversarial attacks remain a credible threat even under severe attacker-defender architecture and dataset mismatches. Notably, the ASR remains non-negligible (~50%) in cross-dataset scenarios, indicating that attackers do not require access to the defender's data distribution to significantly degrade detection performance. The high transferability of PGD highlights the limitations of defenses that focus solely on white-box robustness.

Table 1: Attack effectiveness per DUMB case - ASR (%)

| ASR (%) ↑ | In-dataset | | Cross-dataset | |
|---|---|---|---|---|
| Attack \ DUMB Case | C1 | C3 | C5 | C7 |
| PGD | **99.6** | **89.8** | **65.1** | **66.7** |
| FGSM | 71.7 | 56.5 | 48.4 | 49.2 |
| FPBA | 85.8 | 59.9 | 45.5 | 51.5 |

### 4.2 Defense Validation: Does AT preserve nominal performance?

A defense that sacrifices nominal performance is impractical. First, we ensure that our models are reliable when trained under clean data, before verifying that adversarial training does not harm nominal performance. We train the models on FF++ and evaluate each adversarial training strategy on both in-distribution (FF++) and out-of-distribution (Celeb-DF-V2) test sets. Table 2 presents the AUC for each AT strategy: Base (no AT), retraining on data generated by an attack (FGSM, PGD or FPBA) with the same surrogate, Ens corresponds to retraining on data generated by all attacks (PGD, FGSM, FPBA) using the same surrogate as the model to train, Ens_Surr use all attacks and surrogates for model training.

Table 2: Nominal performance (AUC ↑) per adversarial training strategy

| AT Strategies | Base | PGD | FGSM | FPBA | Ens | Surr | Ens_Surr |
|---|---|---|---|---|---|---|---|
| FF++ | 95.4 | **95.5** | 94.9 | 94.9 | 95.1 | 95.1 | 94.3 |
| Celeb-DF-V2 | 79.8 | 81.7 | 82.5 | 81.9 | 82.1 | 82.3 | **82.9** |

All AT strategies maintained high clean performance, with AUC values exceeding 94% on FF++ and robust performance on Celeb-DF-V2, with AUC ranging from 81.9% to 82.9%. Notably, no catastrophic overfitting was observed, even when employing multi-attack or multi-surrogate training strategies. The ensemble and surrogate-based AT strategies preserved nominal performance, with limited fluctuations compared to the baseline. The preservation of high AUC across all strategies and datasets supports the practical deployment of AT in real-world systems.

**4.3  Ablation - Does Diversity in AT Improve Robustness?**

We hypothesize that increasing the diversity of adversarial examples sources used during training – by incorporating multiple attacks and surrogate models – can enhance robustness, particularly in transfer-based attack scenarios. To test this, we compare the ASR and the attack mitigation rate (AMR) in several AT strategies: single, ensemble and surrogate-based across DUMB cases. We present in Table 3 the adversarial training strategy effectiveness per DUMB case. We present in Table 4 the ASR of AT strategies in leave-on-out scenarios, where we train on two attacks and assess the robustness performance on the last one. As in previous experiments, all results are computed using the ASR equations, which is an average of the number of successful attacks over detectors corresponding to each DUMB case.

Table 3: Adversarial training strategy effectiveness – ASR / AMR (%)

| ASR ↓ / AMR ↑ | In-Dataset | | Cross-Dataset | |
| --- | --- | --- | --- | --- |
| AT Strategy | C1 | C3 | C5 | C7 |
| Base | 85.7 | 68.7 | 53 | 55.8 |
| PGD | 20.2 / 79.7 | 34.2 / 61.9 | 58.4 / 10.3 | 59.9 / 10.2 |
| PGD_surr | **16.1** / 83.8 | **14.4** / 83.9 | 55.7 / **14.4** | 54.1 / **18.9** |
| FGSM | 31.5 / 56.1 | 37.1 / 34.3 | 66.1 / -36.6 | 66.0 / -34.1 |
| FGSM_surr | 31.0 / 56.8 | 25.1 / 55.6 | 58.8 / -21.5 | 55.7 / -13.2 |
| FPBA | 28.6 / 66.6 | 44.3 / 26.0 | **47.5** / -4.4 | 51.7 / -0.4 |
| FPBA_surr | 32.2 / 62.5 | 25.9 / 56.8 | 49.5 / -8.8 | **48.9** / 5.1 |
| Ens | 1.1 / **98.7** | 20.1 / 71.7 | 59.1 / - 11.5 | 60.8 / -0.04 |
| Surr | 26.4 / 69.1 | 21.8 / 69.3 | 54.7 / - 3.1 | 52.9 / 8.9 |
| Ens_Surr | 5.2 / 93.9 | **6.2** / **90.2** | 57.3 / - 8.2 | 57.0 / 2.0 |

**In-distribution robustness gains**. Results in Table 3 reveal that AT with a single strong attack such as PGD yields substantial reductions in ASR, with AMR values indicating robust mitigation (C1 AT PGD reduces ASR from 99.6% to 20.2%). Further gains are observed when increasing the diversity of adversarial examples. Ensemble AT (combining PGD, FGSM and FPBA) achieve the lowest ASR in C1 case with 1.1% ASR, and a strong reduction in C3. This extremely low ASR is the result of a total white box case, where models are tested on samples created by attacks seen in training, and surrogate of the same architecture. Detailed results can be seen in Appendix. Surrogate AT also improves robustness, with PGD_surr and Ens_surr strategies achieving lower ASR compared to their single-attack counterparts, with an exception for FPBA in C1 and C5 configurations. These results confirm that exposing models to a broader spectrum of adversarial perturbations during training enhances their ability to generalize to unseen attacks and transfer scenarios. The improvement is particularly pronounced in cross-model configuration C3, where the attacker and defender share the dataset but differ in model architecture.

Table 4: Leave-one-out multi-attack in adversarial training– ASR (%)

| ASR ↓ | Test Attack | C1 | C3 | C5 | C7 |
|---|---|---|---|---|---|
| PGD_FGSM | FPBA | 35.7 | 39.9 | 64.1 | 66.2 |
| PGD_FPBA | FGSM | **29.7** | **38.9** | **55.5** | **56.5** |
| FGSM_FPBA | PGD | **29.7** | 39.5 | 68.1 | 72.9 |

**Cross-dataset robustness gains.** The benefits of diversity in AT, however, diminish under cross-dataset (C5) and cross-dataset & cross-model (C7) conditions. The lowest ASR value in C7 case corresponds to FPBA_surr, suggesting that leveraging adversarial samples crafted by a deepfake specific attack provides useful information to the models to generalize about unseen attacks. Even Ens_Surr, the most diverse AT strategy yields modest reductions in ASR compared to the baseline and remains above 50% in some cases. This suggests that the robustness conferred by AT is highly sensitive to the alignment between training and evaluation distributions. While ensemble and surrogate-based strategies still outperform single-attack AT in some cases, the AMR gains are smaller, and in certain configurations such as FGSM, FPBA, Ens and Surr, negative AMR values are observed, indicating that AT can inadvertently increase vulnerability. We hypothesise that observed negative AMR values can be attributed to models' overfitting on adversarial artifacts distribution, leading them to be less accurate on perturbations from other distributions.

**Leave-one-out analysis.** Results are presented in Table 4, where ensemble AT is performed with one attack omitted to evaluate robustness performances. Results show that the omission of any single attack from the ensemble leads to a consequent increase in ASR across all DUMB cases. This finding underscores the importance of a comprehensive attack coverage during training. Notably, the omission of FPBA (PGD_FGSM) results in higher ASR in cross-dataset settings (C5: 64.1%, C7: 66.2%) compared to the full ensemble, highlighting the unique threat posed by frequency-based attacks and the necessity of including domain-specific perturbations in the training mix.

**Benefits of surrogate AT.** Surrogate-based AT strategies (e.g., PGD_surr, FGSM_surr, FPBA_surr) demonstrate that leveraging adversarial examples crafted from multiple model architectures can further enhance robustness when compared to their single-attack counterpart (e.g. PGD, FGSM, FPBA), particularly in cross-model and cross-dataset scenarios. However, gains are not uniform: while surrogate diversity helps mitigate overfitting to a single model's vulnerabilities, it cannot fully compensate for distributional shifts between training and evaluation data.

This ablation study findings validate the hypothesis that diversity in adversarial training – both in terms of attack type and surrogate model- improves generalization and robustness in realistic transfer-based attack scenarios. However, the limited effectiveness of AT under cross-dataset conditions reveals a critical limitation: adversarial training alone is insufficient to guarantee robustness against all forms of distribution shift. The occurrence of negative AMR in high-mismatch cases suggests that, without careful evaluation, AT may even introduce new vulnerabilities. While multi-attack and multi-surrogate adversarial training strategies are essential for improving robustness in realistic threat models, additional defenses and rigorous evaluation protocols must complement them.

# 5 Conclusion

This work extends the DUMB/DUMBer methodology to the deepfake detection domain. We introduced a benchmark and experimental framework designed to evaluate adversarial robustness under transfer conditions for the deepfake detection scenario. Our approach explores adversarial transferability across five state-of-the-art deepfake detectors, three attack families, and two datasets (FaceForensics++ and Celeb-DF-V2), mapping results to mismatch cases: C1-C3-C5-C7. We studied three central questions:

**(i) How effective are adversarial attacks when the attacker and defender differ in architecture or dataset?**

Our results show that adversarial attacks remain highly effective even under severe mismatches. Under cross-dataset conditions, transferability decreases but does not vanish, with ASR values around 50%. This confirms that transferability is a practical threat for deployed deepfake detectors.

**(ii) Does adversarial training maintain clean accuracy while improving robustness?**

Yes, AT preserves nominal performance across all strategies (AUC consistently above 94%). However, gains in robustness are uneven: while AT substantially reduces ASR in white-box and gray-box scenarios, its benefits diminish under dataset shift. Negative AMR values in high-mismatch cases highlight that adversarial training can even harm robustness in certain conditions.

**(iii) Does increasing diversity in adversarial training (multiple attacks and surrogates) enhance defense performance?**

Our ablation study illustrates that diversity improves robustness, with AMR improvements between 26% up to 83% in gray-box settings. These strategies also tend to generalize better about unseen attacks, where FPBA AT showed the best result in C7 case and in the leave-one-out experiment when used jointly with PGD. This tends to indicate including task-specific attacks into the AT strengthens defenses. However, AT can also harm defenses, with observed negative AMR under cross-model and cross-dataset scenarios.

Adversarial training is necessary but insufficient in mismatch scenarios. Future works should include other attacks and datasets to provide a clearer understanding of generalization capabilities. Hybrid defenses, adaptive strategies and media compression variations should also be included to close the gap between in-distribution robustness and cross-domain resilience.

# Acknowledgments

The authors express their gratitude to Juan-David DUGAND, Etienne DELATE, Yannick TEGLIA, Ali ZEAMARI for their constructive feedback and support during the preparation of this work. This research did not receive any specific grant from funding agencies in the public sector. The authors declare no competing financial or personal interests that could have influenced the results presented in this study.

# 6 Appendix

## 6.1 Detailed results for adversarial training strategies

We report in Table 5, 6, and 7 the ASR after adversarial training using PGD, FGSM or FPBA. These results illustrate the models' ability to learn from the artifact of one attack, but to not transfer to other attacks. Similarly, we report in Table 8 and 9 the ASR after adversarial training with Ens AT and Ens_Surr AT. This indicates that under a white box setting, with an attack built with a similar architecture, models can learn from the artifacts let by an attack. Similar results can be observed in Table 9, where ASR is low after retraining. However, one can note that the ASR of SPSL for FPBA attack, and SRM for FGSM attacks remain higher, which indicates that using samples crafted by too many surrogates and attacks in training can harm models training.

Table 5: ASR for PGD AT – C1

| ASR ↓ | Xception | UCF | SPSL | SRM | RECCE |
|---|---|---|---|---|---|
| PGD | **0.3** | **0.01** | **0.14** | **0.03** | **0.01** |
| FGSM | 6.96 | 24.1 | 42.7 | 40.2 | 41.4 |
| FPBA | 10.9 | 37.9 | 35.7 | 28.7 | 34.4 |

Table 6: ASR for FGSM AT – C1

| ASR ↓ | Xception | UCF | SPSL | SRM | RECCE |
|---|---|---|---|---|---|
| PGD | 53.8 | 51.8 | 67.6 | 7.9 | 72.0 |
| FGSM | **0.05** | **0.3** | **0.2** | **1.4** | **0.05** |
| FPBA | 35.9 | 41.8 | 68.5 | 45.9 | 24.9 |

Table 7: ASR for FPBA AT – C1

| ASR ↓ | Xception | UCF | SPSL | SRM | RECCE |
|---|---|---|---|---|---|
| PGD | 38.7 | 66.0 | 42.9 | 64.1 | 50.0 |
| FGSM | 25.3 | 26.4 | 13.6 | 48.2 | 46.9 |
| FPBA | **0.03** | **0.4** | **6.7** | **0.01** | **0.03** |

Table 8: Detailed results for Ens AT – C1

| ASR ↓ | Xception | UCF | SPSL | SRM | RECCE |
|---|---|---|---|---|---|
| PGD | **0.14** | **0.11** | **0.11** | **0.17** | **0.08** |
| FGSM | 0.64 | 0.99 | 1.28 | 4.63 | 0.25 |
| FPBA | **0.14** | 0.86 | 7.29 | 0.33 | 0.14 |

Table 9: Detailed results for Ens_Surr AT – C1

| ASR ↓ | Xception | UCF | SPSL | SRM | RECCE |
|---|---|---|---|---|---|
| PGD | 6.77 | **0.81** | **0.05** | **1.75** | **0.27** |
| FGSM | **2.88** | 1.64 | 1.22 | 19.4 | 1.44 |
| FPBA | 2.99 | 6.13 | 21.36 | 9.24 | 2.25 |

## 6.2 ASR equations in C5 and C7 cases

We provide the equations Eq. 4 and Eq. 5 that correspond to the ASR for C5 and C7 cases. In the C5 case, we compute the number of successful attacks for each model $M_i$, attack $A_k$ and dataset $D_2$, corresponding to same model but cross-dataset evaluation (i.e $D_2$ corresponds to Celeb-DF-V2). In the C7 case, we compute the number of successful attacks on target model $M_i$, using adversarial samples from attack $A_k$ created with the surrogate $M_s$ on dataset $D_2$.

$$\text{ASR}_{C5}^{(D_2)} = \frac{1}{|\mathcal{M}| \cdot |\mathcal{A}|} \sum_{A_k \in \mathcal{A}} \sum_{M_i \in \mathcal{M}} \frac{N_{\text{success}}^{(M_i, A_k, D_2)}}{N_{\text{total}}^{(M_i, A_k, D_2)}} \qquad (Eq. 4)$$

$$\text{ASR}_{C7}^{(D_2)} = \frac{1}{|\mathcal{M}|(|\mathcal{M}|-1) \cdot |\mathcal{A}|} \sum_{A_k \in \mathcal{A}} \sum_{M_t \in \mathcal{M}} \sum_{M_s \in \mathcal{M}, M_s \neq M_t} \frac{N_{\text{success}}^{(M_t \leftarrow M_s, A_k, D_2)}}{N_{\text{total}}^{(M_t \leftarrow M_s, A_k, D_2)}} \quad (Eq.5)$$